\documentclass[10pt,conference,a4paper]{IEEEtran}
\IEEEoverridecommandlockouts
\usepackage{cite}
\usepackage{amsmath,amssymb,amsfonts}
\usepackage{algorithmic}
\usepackage{graphicx}
\usepackage{textcomp}
\usepackage{xcolor}
\usepackage[colorlinks,linkcolor=red]{hyperref}
\def\BibTeX{{\rm B\kern-.05em{\sc i\kern-.025em b}\kern-.08em
    T\kern-.1667em\lower.7ex\hbox{E}\kern-.125emX}}
\begin{document}

\title{Galaxy Image Translation with Semi-supervised Noise-reconstructed Generative Adversarial Networks\\
\thanks{This project has received funding from the European Union's Horizon 2020 research and innovation programme under the Marie Skłodowska-Curie grant agreement No713750. Also, it has been carried out with the financial support of the Regional Council of Provence-Alpes-C\^{o}te d'Azur and with the financial support of the A*MIDEX (n° ANR-11-IDEX-0001-02), funded by the Investissements d'Avenir project funded by the French Government, managed by the French National Research Agency (ANR).}}

\author{\IEEEauthorblockN{1\textsuperscript{st} Qiufan Lin}
\IEEEauthorblockA{
\textit{Aix Marseille Univ., CNRS/IN2P3, CPPM}\\
Marseille, France \\
lin@cppm.in2p3.fr}
\and
\IEEEauthorblockN{2\textsuperscript{nd} Dominique Fouchez}
\IEEEauthorblockA{
\textit{Aix Marseille Univ., CNRS/IN2P3, CPPM}\\
Marseille, France \\
fouchez@cppm.in2p3.fr}
\and
\IEEEauthorblockN{3\textsuperscript{rd} J\'{e}r\^{o}me Pasquet}
\IEEEauthorblockA{\textit{UMR TETIS, Univ. Montpellier} \\
\textit{AgroParisTech, Cirad, CNRS, Irstea}\\
Montpellier, France \\
jerome.pasquet@univ-montp3.fr}
}

\maketitle

\begin{abstract}
Image-to-image translation with Deep Learning neural networks, particularly with Generative Adversarial Networks (GANs), is one of the most powerful methods for simulating astronomical images. However, current work is limited to utilizing paired images with supervised translation, and there has been rare discussion on reconstructing noise background that encodes instrumental and observational effects. These limitations might be harmful for subsequent scientific applications in astrophysics. Therefore, we aim to develop methods for using unpaired images and preserving noise characteristics in image translation. In this work, we propose a two-way image translation model using GANs that exploits both paired and unpaired images in a semi-supervised manner, and introduce a noise emulating module that is able to learn and reconstruct noise characterized by high-frequency features. By experimenting on multi-band galaxy images from the Sloan Digital Sky Survey (SDSS) and the Canada France Hawaii Telescope Legacy Survey (CFHT), we show that our method recovers global and local properties effectively and outperforms benchmark image translation models. To our best knowledge, this work is the first attempt to apply semi-supervised methods and noise reconstruction techniques in astrophysical studies.
\end{abstract}


\section{Introduction}

Simulating realistic astronomical images is an important but hard task in astrophysics. Other than real observational data, astronomers utilize simulated images in a variety scientific studies, ranging from analyzing single celestial objects (e.g., transients, stars, galaxies, etc.) to probing the evolution of the universe (e.g., weak gravitational lensing, large-scale structures, supernova cosmology, etc.). Unlike traditional non-Deep-Learning simulation methods, generative models built on Generative Adversarial Networks (GANs) have shown promises in producing high-fidelity images in an efficient fashion without imposing theoretical modeling assumptions (e.g., \cite{CosmoGAN}, \cite{1811.03081}, \cite{RadioGAN}).
While generative models using random seeds as input tend to ``invent'' new content, image-to-image translation is able to generate new images retaining the content learned from a source domain. This approach has its merits when we attempt to augment data for real tasks for a target sky survey based on other existing surveys differing in instrumental and observational effects. In this regard, our work specifically focuses on making image simulations via image-to-image translation.

A straightforward way of making image translation would be to use image pairs from two surveys that contain the same objects, each serving as the ``ground truth'' for the other in a supervised manner. However, due to limited overlapping sky coverage, we are usually lacking such paired data to train large-scale neural networks, while there are always sufficient unpaired data to use. Moreover, models trained with paired data alone might not be able to generalize well over the distribution of the unpaired data. As current image translation work in astrophysics is limited to using paired data, we aim to develop an unsupervised or semi-supervised translation method for exploiting unpaired data.

Another major part in our work is noise reconstruction in image translation. One image can be decomposed into useful signal and noise. We define the useful signal (or the non-noise part) as the true signal from the object convolved with the Point Spread Function (PSF), predominantly governed by atmospheric blurring for ground-based telescopes and usually characterized as low-frequency features.
Background noise is the other important component of an image. Because of its existence and variations in nature, even the same celestial objects may have dissimilar appearances on images from different surveys. Noise could come from divergent sources, including shot noise from the object itself and the sky background under the object, thermal radiation inside the detector, or random loss and gain of electrons during the CCD read-out process. As noise encodes systematic effects of a survey, models developed upon training data with particular noise characteristics might fail on previously unseen data with different noise properties. In addition, noise is an unavoidable element in the simulation of realistic data for a survey. Despite its importance, noise is usually overlooked as it is hard to learn and preserve due to its high-frequency nature. We are cautious that noise would bias signal recovery and jeopardize subsequent analyses if not treated properly. We note that CycleGAN-like image translation implementations are already capable of reconstructing non-noise patterns, but they lack ingredients to generate noise. There have been studies discussing noise modeling for natural image processing (e.g., \cite{NoiseA}, \cite{NoiseB}), but they are restricted to pre-defined and artificially generated noise patterns. Unlike other studies that focus on removing noise from images based on noise modeling or simulations, we attempt to preserve noise information from real images and reconstruct noise in image translation.

The contributions of this work are as follows.
\begin{enumerate}
	\item We develop a two-way one-to-one mapping model\footnote{Code is available at \href{https://github.com/QiufanLin/ImageTranslation}{https://github.com/QiufanLin/ImageTranslation}.} for galaxy image translation using Generative Adversarial Networks (GANs). Our semi-supervised training scheme makes use of not only the unpaired data representative of the distribution of the full dataset, but also the paired data that ensures precise calibration of the target domain.
	\item We achieve to reconstruct noise by introducing noise emulating modules and discriminators that concentrate on high-frequency features. Though developed for astronomical imaging analysis, this technique might also be beneficial for image processing tasks in other fields.
\end{enumerate}

\section{Related Work}

\subsection{Generative Adversarial Models in Astrophysics}

Recently, the GAN networks \cite{GAN} have become increasingly popular in astrophysical studies. The \textit{vanilla} GAN networks set up a minmax game between a Generator and a Discriminator, taking a random noise seed to produce fake data --- such as images \cite{DCGAN} --- as expected to be indistinguishable from the target domain. This has been applied as generative models in image simulations such as \cite{CosmoGAN} and \cite{1811.03081}. \cite{CosmoGAN} achieved to simulate weak lensing convergence maps with high statistical confidence. \cite{1811.03081} developed a chained method to produce galaxy images with high resolution based on StackGAN \cite{StackGAN}, by first producing low-resolution images from random seeds and then upsampling low-resolution images.

The GAN method has also been implemented in studies relevant to image translation. For example, \cite{1812.05781} used image translation method to capture the underlying noise field from a noisy weak lensing convergence map. Similarly, \cite{1702.00403} trained a model to recover clean images from degraded images with bad astronomical seeing and high noise. \cite{1810.10098} proposed a branched GAN network to deblend overlapping galaxies. We note that all of these studies trained the networks with simulated data, from which the ground truth information for supervised translation is accessible. \cite{RadioGAN} made image translation using real data from two radio surveys with different resolutions and brightness sensitivities, yet they only exploit paired image cutouts.

\subsection{Two-way Adversarial Networks}

Two-way GAN networks in general hold two minmax games in parallel, having two Generators and two Discriminators, which are built to make connections across two domains. Though not specifically referred to as two-way translation, the idea of connecting domains has existed in early work such as ALI \cite{ALI} and BiGAN \cite{BiGAN}, where an inverse mapping is made to recover the random seed inputted for generating fake data. Similarly, InfoGAN \cite{InfoGAN} is trained to maximize the mutual information between generated data and a second seed used as a conditional code \cite{CGAN}. CoGAN \cite{CoGAN} produces images from a joint distribution learned from different domains.

The concept of two-way translation became clear when CycleGAN \cite{CycleGAN}, as an extension of pix2pix \cite{pix2pix}, applied a cycle-consistency loss to ensure one-to-one mapping between two domains. It was proposed to make use of unpaired data for training. Similar ideas can also be found in other work. StarGAN \cite{StarGAN} makes multi-domain translation controlled by domain-specified labels. Translation between each pair of domains is constrained by a cycle-consistency loss. DualGAN \cite{DualGAN} adopts the same cycle-consistency constraint, except that the adversarial loss follows Wasserstein GAN \cite{WGAN}. DiscoGAN \cite{DiscoGAN} explores a few forms of cycle-consistency loss. XGAN \cite{XGAN} applies semantic consistency to embedded features. While CycleGAN achieves approximately deterministic mappings between domains, Augmented CycleGAN \cite{AugCGAN} extends the idea of CycleGAN by introducing stochastic many-to-many mappings. TraVeLGAN \cite{TraVeLGAN} applies a siamese network to preserve high-level intra-domain semantics to get rid of the commonly-used cycle-consistency loss.

To our best knowledge, unsupervised or semi-supervised methods have not yet been investigated in any research of astrophysics prior to our work, though they have already been heavily explored in the field of computer science. In addition, previous work predominantly focuses on de-noising techniques, whereas there has been rare discussion on noise reconstruction.

\begin{figure}[ht]
\begin{center}
\centerline{\includegraphics[width=\columnwidth]{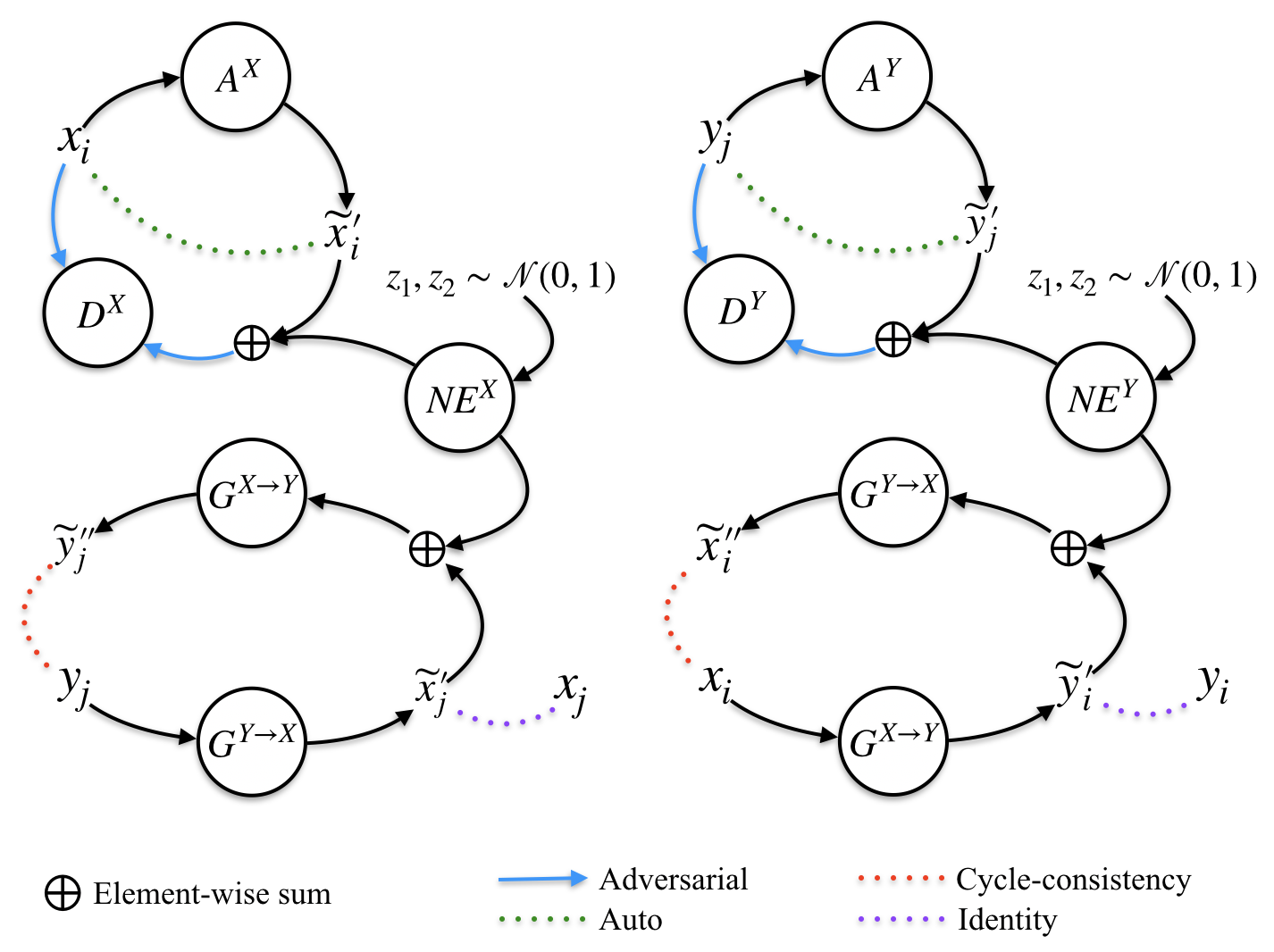}}
\caption{Framework of our two-step training scheme. \textbf{Step One:} (i) update the Autoencoders $A^X$, $A^Y$ with the original images $x$, $y$ from the two domains $X$, $Y$, respectively; (ii) adversarially update the Noise Emulators $NE^X$, $NE^Y$ and the Discriminators $D^X$, $D^Y$ while keeping $A^X$, $A^Y$ (thus the generated non-noise images $\widetilde{x}$, $\widetilde{y}$) fixed and taking Gaussian random seeds $z_1$, $z_2$ as inputs to $NE^X$, $NE^Y$ to produce noise. \textbf{Step Two:} update the Generators $G^{X \rightarrow Y}$, $G^{Y \rightarrow X}$, using noise produced by $NE^X$, $NE^Y$.}
\label{fig:graph}
\end{center}
\end{figure}

\begin{figure*}
\begin{center}
\centerline{\includegraphics[width=1.0\linewidth]{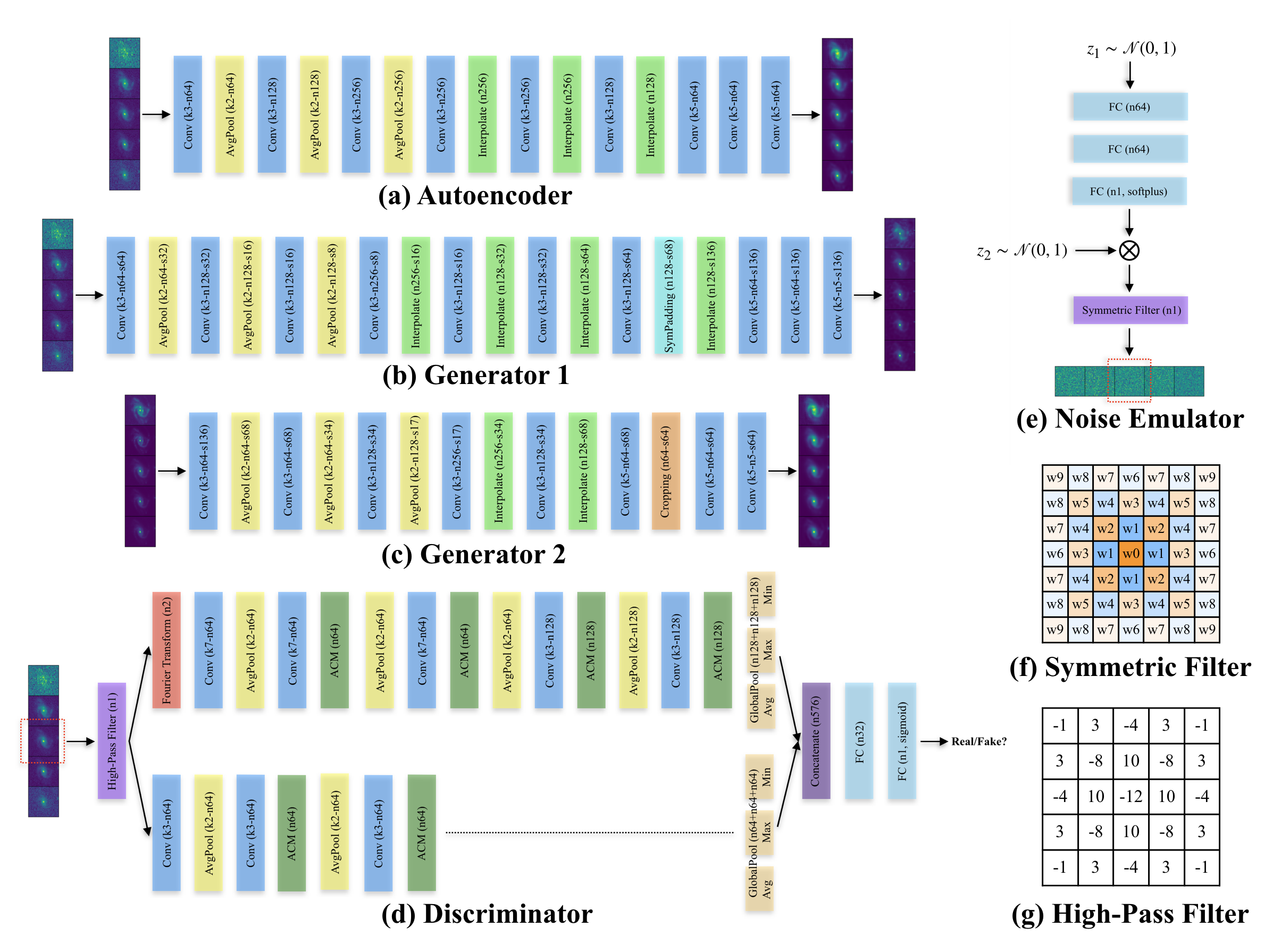}}
\caption{The architecture networks in our model. ``Conv'', ``AvgPool'' and ``FC'' refer to Convolutional layers, Average Pooling layers and Fully-Connected layers, respectively. ``GlobalPool'' refers to Global Pooling layers, divided into Average Pooling, Maximum Pooling and Minimum Pooling. ``Interpolate'' and ``SymPadding'' refer to the Nearest Neighbor Interpolation and the Symmetric Padding, used as upsampling operations. The reversed downsampling operation is ``Cropping''. ``ACM'' refers to the Attention Complementary Modules implemented by~\cite{ACNet}. The labels ``k'', ``n'' and ``s'' refer to the size of the convolutional filter, the number of output channels and the size of the output feature map or image. The output from the 2D Fourier Transform is separate as real and imagery parts in two channels.
We adopt Leaky ReLUs with a slope of 0.2 unless otherwise noted. \textbf{(a)} Architecture of the Autoencoders. \textbf{(b)} Architecture of the Generator translating SDSS images to non-noise CFHT images. \textbf{(c)} Architecture of the Generator translating CFHT images to non-noise SDSS images. The size of each output feature map or image is specified. \textbf{(d)} Architecture of the Discriminators for each passband. \textbf{(e)} Architecture of the Noise Emulators for each passband. $z_1$ and $z_2$ denote the injected Gaussian random seeds. ``$\bigotimes$'' denotes element-wise product. \textbf{(f)} The 7$\times$7 zero-bias symmetric filter applied in the Noise Emulators. \textbf{(g)} The high-pass filter applied in the Discriminators.}
\label{fig:architecture}
\end{center}
\end{figure*}

\section{Our Model} \label{sec:model}

Our two-way image-to-image translation model connects two domains $X$ and $Y$ via two mapping functions, $F^{X \rightarrow Y}$ and $F^{Y \rightarrow X}$. Each mapping function is composed of two components --- a Generator that decouples noise from the input image and generates the non-noise part of the target image, and a Noise Emulator that reconstructs a noise background associated with the target domain. The output target image is the sum of these two parts ($F^{X \rightarrow Y}=G^{X \rightarrow Y}+NE^Y$, $F^{Y \rightarrow X}=G^{Y \rightarrow X}+NE^X$). The amplitude of the noise background changes from image to image, whose distribution is what we want to learn and sample via adversarial training with two Discriminators.

If we directly train the Generators and the Noise Emulators with two Discriminators like other two-way translation methods, we would encounter difficulties in regenerating well-behaved noise. Noise has to be learned together with the non-noise part of an image, as they cannot be split \textit{a priori}. For translation from a high-resolution domain to a low-resolution domain, the Generator is fed with high-resolution information that is redundant for generating low-resolution images, and the gradients propagated from the Discriminator would force the Generator to produce high-frequency fluctuations that mimic the behavior of noise but actually hinder the training of the Noise Emulator. On the other side, noise reconstruction is unlikely to succeed for translation from a low-resolution domain to a high-resolution domain, since the Discriminator would stick with detailed yet non-noise features that can never be regenerated by the Generator with low-resolution images as input.

Therefore, we speculate that noise reconstruction would only be achievable through translation within the same domain, i.e., via the use of Autoencoders. Moreover, having them trained separately from the Discriminators can avert noise-like patterns harmful for the Noise Emulators. Adopting two independent Autoencoders as auxiliary components is a trade-off between the two aforementioned challenging situations, which enables the Noise Emulators to be properly optimized.

\subsection{Training Scheme and Objective Functions} 

Having these considerations, we propose a two-step training scheme (Figure~\ref{fig:graph}). In Step One, the Noise Emulators $NE^X$, $NE^Y$ are first trained with two Discriminators $D^X$, $D^Y$ and two Autoencoders $A^X$, $A^Y$. The noise generated by the Noise Emulators is added to the non-noise part generated by the Autoencoders and fed as input into the Discriminators.
For each iteration of the training, we update the Autoencoders independently from the other two components, then update the Discriminators and the Noise Emulators via adversarial training while keeping the Autoencoders fixed, i.e., we minimize the auto and adversarial losses alternately.

\begin{scriptsize}
\begin{equation}
\begin{aligned}
\mathcal{L}_{auto}(A^X,A^Y)
&= \mathbb{E}_{x \sim X} [\| A^X(x) - x \|_2] \\
&+ \mathbb{E}_{y \sim Y} [\| A^Y(y) - y \|_2]
\label{eq:A}
\end{aligned}
\end{equation}
\begin{equation}
\begin{aligned}
\mathcal{L}_{adv}(D^X,D^Y)
&= \mathbb{E}_{x \sim X} [-\sum_{p=1}^{N_p} (\log D^X_p(x) + \log (1 - D^X_p(F^X(x))))] \\
&+ \mathbb{E}_{y \sim Y} [-\sum_{p=1}^{N_p} (\log D^Y_p(y) + \log (1 - D^Y_p(F^Y(y))))]
\end{aligned}
\label{eq:D}
\end{equation}
\begin{equation}
\begin{aligned}
\mathcal{L}_{adv}(NE^X,NE^Y) &= \mathbb{E}_{x \sim X} [- \sum_{p=1}^{N_p} (\log D^X_p(F^X(x)))] \\
&+ \mathbb{E}_{y \sim Y} [- \sum_{p=1}^{N_p} (\log D^Y_p(F^Y(y)))]
\end{aligned}
\label{eq:F}
\end{equation}
\end{scriptsize}$X$ and $Y$ denote two domains. $D^X_p$ and $D^Y_p$ denote the Sub-Discriminators for the passband $p$ (discussed in Section~\ref{dis}), running over the total number of passbands $N_p$. $F^X=A^X+NE^X$ and $F^Y=A^Y+NE^Y$ denote the self-mapping functions combining the Autoencoders $A^X$, $A^Y$ and the Noise Emulators $NE^X$, $NE^Y$.

In Step Two, both the Discriminators and the Autoencoders are discarded. We add the noise reconstructed from the trained Noise Emulators to the non-noise output of the Generators $G^{X \rightarrow Y}$, $G^{Y \rightarrow X}$ and train the Generators in the form of a cycle. The identity and cycle-consistency losses are minimized with paired and unpaired data respectively.

\begin{scriptsize}
\begin{equation}
\begin{aligned}
\mathcal{L}_{id}(G^{X \rightarrow Y},G^{Y \rightarrow X})
&= \mathbb{E}_{(x,y) \sim (X,Y)_{pair}} [\| G^{X \rightarrow Y}(x) - y \|_2] \\
&+ \mathbb{E}_{(x,y) \sim (X,Y)_{pair}} [\| G^{Y \rightarrow X}(y) - x \|_2]
\label{eq:id}
\end{aligned}
\end{equation}
\begin{equation}
\begin{aligned}
\mathcal{L}_{cyc}(G^{X \rightarrow Y},G^{Y \rightarrow X})
&= \mathbb{E}_{x \sim X_{unpair}} [\| G^{Y \rightarrow X}(F^{X \rightarrow Y}(x)) - x \|_2] \\
&+ \mathbb{E}_{y \sim Y_{unpair}} [\| G^{X \rightarrow Y}(F^{Y \rightarrow X}(y)) - y \|_2] \\
\end{aligned}
\label{eq:cyc}
\end{equation}
\end{scriptsize}$F^{X \rightarrow Y}=G^{X \rightarrow Y}+NE^Y$ and $F^{Y \rightarrow X}=G^{Y \rightarrow X}+NE^X$ denote the cross-domain mappings combining the Generators $G^{X \rightarrow Y}$, $G^{Y \rightarrow X}$ and the Noise Emulators $NE^X$, $NE^Y$.

In addition to the loss functions discussed above, one could add extra losses or auxiliary networks so as to put an additional constraint on the information critical to subsequent tasks (e.g., galaxy type classification). We leave this investigation to future analysis, as our goal in this work is only to recover broad galaxy shapes by the content losses (Eq.~\ref{eq:id} and Eq.~\ref{eq:cyc}).

\subsection{Autoencoders}

The architecture of the Autoencoders is shown in Figure~\ref{fig:architecture}(a). We observe in our experiments that applying three Average Pooling layers is sufficient to smooth images of the typical medium-level noise in our study. In order to avoid producing grid patterns, we upsample images using the Nearest Neighbor Interpolation rather than the Deconvolutional layers or the Pixel Shuffle units (\cite{Subpixel,SRGAN}).

\subsection{Generators}

The Generators (Figures~\ref{fig:architecture}(b) and Figures~\ref{fig:architecture}(c)) are used for cross-domain translation, and can be regarded as variants of the Autoencoders. We use the Symmetric Padding/Cropping as additional upsampling/downsampling operations to adjust the image size with the target domain.

\subsection{Noise Emulators}

The Noise Emulators are built upon the following observations and assumptions:
\textbf{(1)} Noise behaves as fluctuations discretized on pixels. There may be correlations among adjacent pixels.
\textbf{(2)} The noise pattern is translationally and rotationally invariant over the relatively small spatial scale of the images used in our work. 
\textbf{(3)} While independent of signal, noise is subject to the properties of a survey and its amplitude may vary due to varying observational or instrumental conditions.
\textbf{(4)} While signal is shared among passbands, we assume that the noise properties associated with one passband from a survey are independent of the other passbands as well as the other survey. Despite this assumption, we are cautious that noise amplitudes among different passbands might be correlated due to certain observation strategies. However, as the noise amplitude for each passband is to be randomly sampled, we do not take this correlation into account.

The architecture of the Noise Emulators is shown in Figure~\ref{fig:architecture}(e). Each channel is independent but shares the same architecture, corresponding to a passband. To generate noise, a random number $z_1$ is sampled from the standard Gaussian distribution, which is transformed into a scalar that controls the noise amplitude; a 2D Gaussian random seed $z_2$ is sampled as a noise map; finally, the noise map is multiplied with the amplitude and convolved with a 7$\times$7 zero-bias symmetric filter (i.e., with shared weights at opposite positions w.r.t the center, Figure~\ref{fig:architecture}(f)) that introduces short-scale correlations.
Once trained, we can use the Noise Emulators to regenerate noise by sampling $z_1$ and $z_2$ for each passband.

\subsection{Discriminators} \label{dis}

Since noise is treated independently among different passbands, we take individual channels as Sub-Discriminators, each corresponding to a passband (Figure~\ref{fig:architecture}(d)). For each channel, we first apply a high-pass filter (Figure~\ref{fig:architecture}(g)) to the input, so that noise is more emphasized relative to low-frequency features. The information flow is then split into two branches, one of which passes through a 2D Fourier Transform module. The output real and imagery parts of the Fourier Transform are concatenated. The ability of discriminating high-frequency features is improved via combining the pixel space and the Fourier space. Furthermore, the feature maps are enhanced by the Attention Complementary Modules (ACMs) proposed by~\cite{ACNet}. After concatenating the global average, maximum and minimum of each feature map in each branch, the Sub-Discriminator outputs a probability indicating how likely the input in the passband is real.

\begin{figure*}
\begin{center}
\centerline{\includegraphics[width=0.9\linewidth]{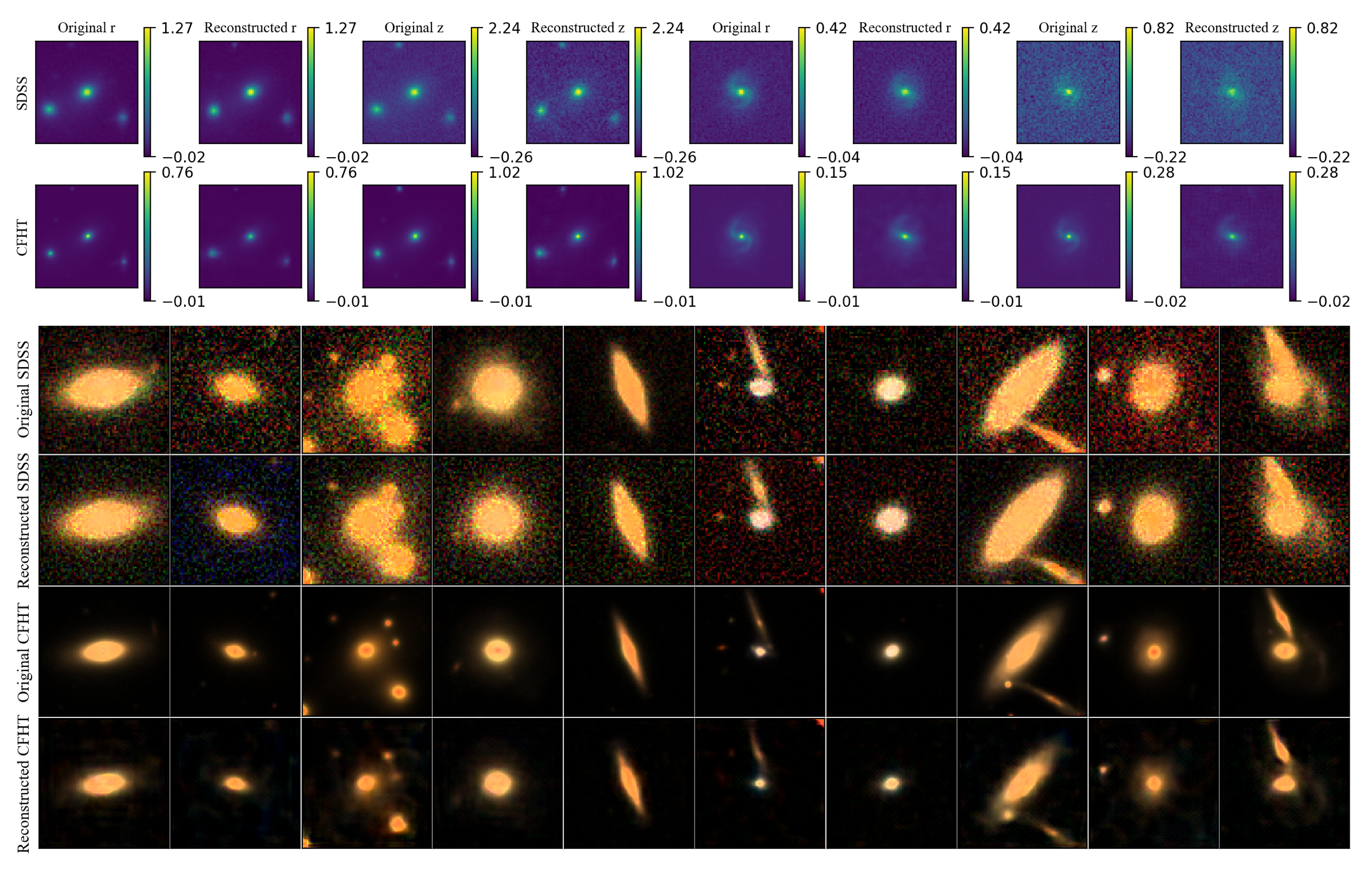}}
\caption{Results of image translation obtained from our model. The upper panel shows reconstructed and original image pairs of two galaxies in $rz$ bands. Negative flux is due to background subtraction in image pre-processing. The lower panel shows a few more examples in RGB format created by the method introduced in~\cite{Lupton_2004}. Note that the noise amplitude for each reconstructed image is randomly sampled from the corresponding Noise Emulator, thus it does not necessarily match the noise amplitude of the original counterpart.}
\label{fig:images}
\end{center}
\end{figure*}

\begin{figure}[ht]
\begin{center}
\centerline{\includegraphics[width=1.0\columnwidth]{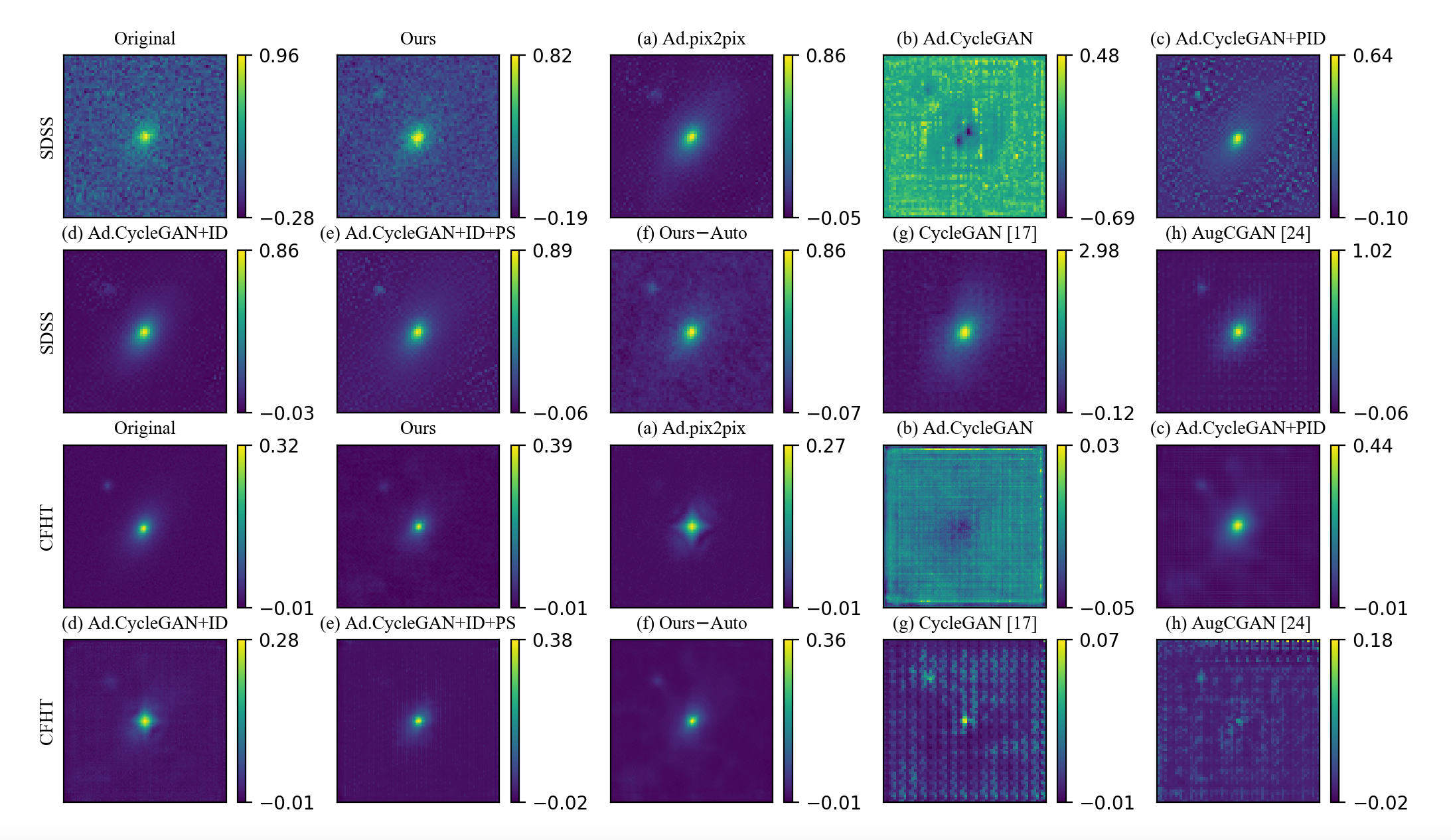}}
\caption{The $z$-band image pairs of a galaxy obtained from variant cases. \textbf{(a)} One-way translation using our networks similar to pix2pix \cite{pix2pix} (``Adapted pix2pix''). \textbf{(b)} Two-way translation using our networks similar to CycleGAN \cite{CycleGAN} (``Adapted CycleGAN''). \textbf{(c)} Adapted CycleGAN with the $pseudo$-identity loss (PID). \textbf{(d)} Adapted CycleGAN with the identity loss (ID). \textbf{(e)} Adapted CycleGAN with the identity loss (ID) and the Pixel Shuffle units (PS). \textbf{(f)} Two-way translation using our model except the Autoencoders. \textbf{(g)} Two-way translation using CycleGAN implemented in \cite{CycleGAN}. \textbf{(h)} Two-way translation using Augmented CycleGAN implemented in \cite{AugCGAN}. More details can be found in the text.}
\label{fig:comparisons}
\end{center}
\end{figure}

\section{Experiments}

\subsection{Data}

The two domains $X$, $Y$ in our experiments are galaxy images from two surveys --- the Sloan Digital Sky Survey (SDSS) and the Canada France Hawaii Telescope Legacy Survey (CFHT), respectively. We take the SDSS dataset retrieved by~\cite{Pasquet_2018} from SDSS Data Release 12 \cite{SDSS}, consisting of 659,821 cutout images of size 64$\times$64 pixels over five photometric passbands ($ugriz$). Each image contains a galaxy at the center. The CFHT cutout sample is created from the CFHT wide field observations W1, W2, W3 and W4 \cite{CFHT}, consisting of 130,093 cutout images with $ugriz$ passbands as well. Due to a different resolution, we choose the CFHT cutout size to be 136$\times$136 pixels so that both SDSS and CFHT cutout images span over the same angular scale. As will be used in our experiments, we also create a sample of CFHT images of size 64$\times$64 pixels as SDSS images by regridding original CFHT images with the Bilinear Interpolation.

We use \texttt{ra} and \texttt{dec} -- the sky coordinate information --- to cross-match images in the two samples and identify 5,057 ones that contain same galaxies and have aligned peak positions (i.e., paired images). Of these images, we take 2,557 as the paired training sample and 2,500 as the test sample. The remaining 654,764 SDSS images and 125,036 CFHT images are used as the unpaired training sample.

\subsection{Experiment Details}

We consider a few variants of our proposed model as ablation analysis and look for substitutes for the identity loss (Eq.~\ref{eq:id}) and the Noise Emulators. To regulate the intermediate output, one way is to use the following $pseudo$-identity loss.

\begin{scriptsize}
\begin{equation}
\begin{aligned}
\mathcal{L}_{pseudo-id}(G^{X \rightarrow Y},G^{Y \rightarrow X}) &= \mathbb{E}_{x \sim X} [\| G^{X \rightarrow Y}(x) - x \|_2] \\
&+ \mathbb{E}_{y \sim Y} [\| G^{Y \rightarrow X}(y) - y \|_2]
\end{aligned}
\label{eq:pseudoid}
\end{equation}
\end{scriptsize}This is different from Eq.~\ref{eq:id} in our work. It acts as a restriction on cross-domain variation with unpaired data rather than a precise identity mapping with paired data. Regarding noise reconstruction, an alternative would be to make use of the Pixel Shuffle units for upsampling, the key element to generating super-resolution images (\cite{Subpixel,SRGAN}). Moreover, CycleGAN \cite{CycleGAN} and Augmented CycleGAN \cite{AugCGAN} are examples of the benchmark translation methods using unpaired images. We are also interested in checking their applicability in our study.

The variant cases we analyze are summarized below: \textbf{(a) Ad.pix2pix:} One-way translation using our networks with the identity loss but not the Autoencoders or the Noise Emulators, similar to pix2pix \cite{pix2pix} (denoted with ``Adapted pix2pix''). \textbf{(b) Ad.CycleGAN:} Two-way translation using our networks with the cycle-consistency loss but not the Autoencoders or the Noise Emulators, similar to CycleGAN \cite{CycleGAN} (denoted with ``Adapted CycleGAN''). \textbf{(c) Ad.CycleGAN+PID:} Same as Case (b), except adding the $pseudo$-identity loss (PID). \textbf{(d) Ad.CycleGAN+ID:} Same as Case (b), except adding the identity loss (ID). \textbf{(e) Ad.CycleGAN+ID+PS:} Same as Case (d), except using the Pixel Shuffle units (PS) in the Generators. \textbf{(f) Ours--Auto:} Same as our model, except trained in one step without the Autoencoders. \textbf{(g) CycleGAN:} Two-way translation with only the cycle-consistency loss, using the 6-residual-block CycleGAN architecture as presented in~\cite{CycleGAN}. \textbf{(h) AugCGAN:} The semi-supervised two-way translation setting of Augmented CycleGAN as presented in~\cite{AugCGAN}, based on the CycleGAN architecture with the identity loss, having random seeds injected to the Generators to enable stochastic mappings.

For all of these cases, the training is completed in one step, in which the Autoencoders are not used, and the images produced by the Generators are inputted to the Discriminators. In Case (f), only the content losses are used to update the Generators (i.e., the adversarial loss is applied to the Noise Emulators rather than the Generators, similar to the training step one for our model); while in the remaining cases without the Noise Emulators, the content losses are multiplied by 1,000 and added to the adversarial loss to update the Generators.

We run 60,000 update iterations for each of these cases and the two steps for our model. In an iteration, we randomly select a mini-batch of 24 unpaired SDSS images and 24 unpaired CFHT images, as well as 24 image pairs (i.e., 96 images in total), with random flipping and rotation by 90 deg steps. The regridded 64$\times$64 CFHT images are used in Cases (g) and (h). The learning rate starts with $10^{-4}$ and is reduced by a factor of 5 every 20,000 iterations. We adopt the default implementation of Adam Optimizer \cite{Adam}, and perform gradient clipping at a ratio of 5 to the gradient norm. Using an Intel(R) Core(TM) i9-7920X CPU and a Titan V GPU, roughly 30 hours is required to complete the two training steps (120,000 iterations in total).

\subsection{Results}

The images reconstructed by our translation model show good quality in visual comparison with the paired real images (Figure~\ref{fig:images}). We are able to regenerate well-behaved noise background that significantly improves the fidelity of reconstructed images. Moreover, our paired training data is sparse yet sufficient to convey the knowledge of the ``ground truth'' identity mapping and calibrate the broad galaxy shapes with good precision. The major deficiency of our method would be the loss of non-noise structures at small spatial scales (e.g., small spots or faint spiral arms) for the reconstructed CFHT images. We note that this is partly due to the compromise of using the Nearest Neighbor Interpolation, and also a common challenge for image translation, as it is difficult for the networks to ``extrapolate'' such detailed information that does not exist on the low-resolution SDSS images.

In contrast to our model, there are limitations in the variant cases (Figure~\ref{fig:comparisons}). In Case (a), while the one-way translation using paired images is capable of conveying identity information, paired data alone cannot represent the full sample distribution dominated by unpaired data. In Cases (b) and (g), although the cycle consistency is guaranteed, the intermediate output from the half-cycle has large variation due to the absence of paired data. The $pseudo$-identity loss in Case (c) helps constrain the intermediate output given the similarity between two domains, whereas a gap still exists from the ``ground truth'' paired image counterparts. This gap cannot be corrected unless the true identity information from paired data is exploited. 
In Cases (a) --- (e) and (g), as no randomness has been introduced to regenerate noise, there exist regular fluctuations mimicking the behavior of noise. In Case (h), the injected seeds might help maintain some level of stochasticity, but fail to ensure the shape reconstruction or recover the correct spatial noise patterns.

\begin{figure}[ht]
\begin{center}
\centerline{\includegraphics[width=\columnwidth]{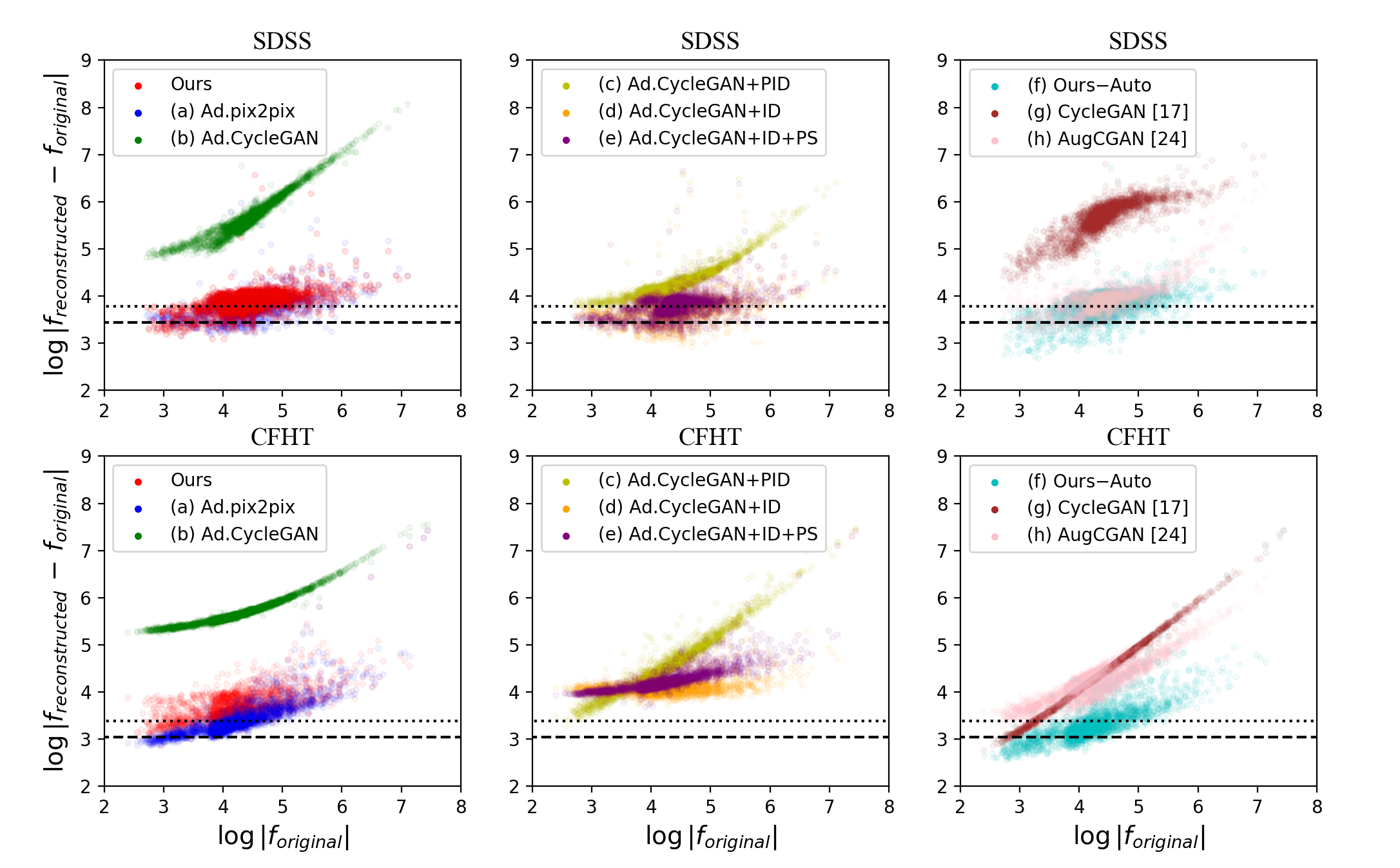}}
\caption{Global shape reconstruction evaluated on spatial flux distributions. Results with the $r$-band images are shown for our model and Cases (a) --- (h) (defined in the text). Both axes display the summed absolute pixel-wise fluxes in the logarithmic scale. The black dotted and dashed lines indicate the flux differences corresponding to the optimal global reconstruction with and without noise reconstruction, respectively (i.e., $\sqrt{2} \times \sigma$ and $\sigma$, where $\sigma$ is the median summed noise amplitude estimated using the Noise Emulators).}
\label{fig:global}
\end{center}
\end{figure}

\begin{figure}[ht]
\begin{center}
\centerline{\includegraphics[width=1.0\columnwidth]{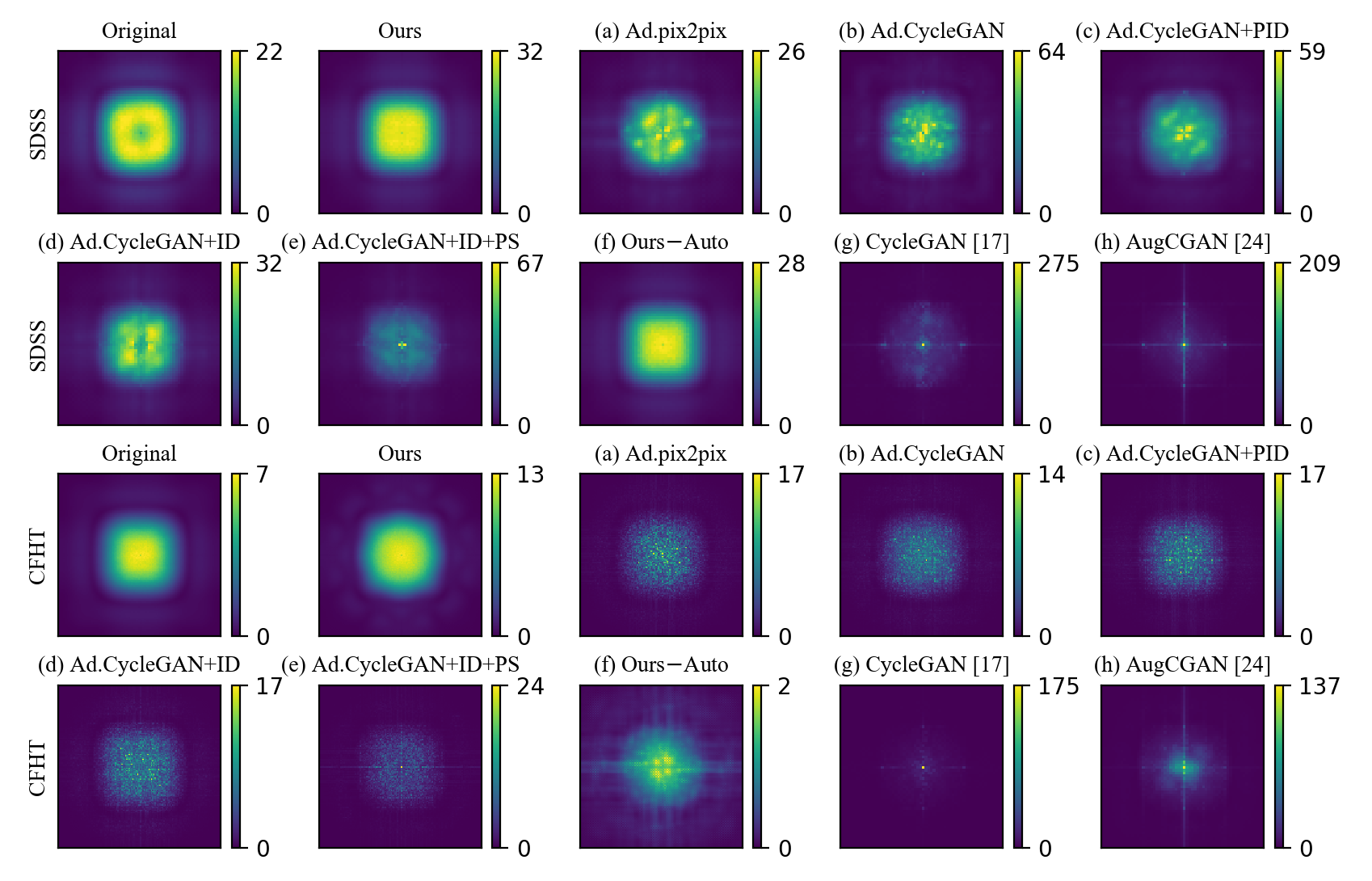}}
\caption{Local fluctuation patterns displayed in the Fourier space. Each panel shows the stacked Fourier amplitude map for the original $r$-band images and those with our model and Cases (a) --- (h) (defined in the text). The CFHT images in Cases (g) and (h) have a modified image resolution due to regridding. The centers correspond to the highest frequency.}
\label{fig:ft}
\end{center}
\end{figure}

\subsection{Evaluation} \label{sec:evaluation}

We define metrics to evaluate the reconstruction of global galaxy shapes and local fluctuation patterns. The evaluation is made on 2,500 image pairs from the test sample and the 5,000 corresponding reconstructed images. Since we generate images with noise of random amplitude, metrics such as SSIM or PSNR are incapable of evaluating image quality in our work. Quantitative evaluation has to be made using metrics specific to real tasks in astrophysics.

\subsubsection{Global Shape Reconstruction}

To evaluate the global shape reconstruction of an image, we treat each of its passbands as a 2D spatial flux distribution on the grid, and sum up the absolute pixel-wise differences between the image and its original counterpart. Figure~\ref{fig:global} shows the $r$-band flux difference as a function of the original flux for all the cases. The other passbands exhibit similar trends. 
The groups of dots produced by all the methods with the identity loss (including our model and Cases (a), (d), (e), (f), (h)) are on average lower than the remaining groups, suggesting a better constraint. As a result of noise, the flux difference cannot reach zero even using a perfect model to recover global shapes. Compared to our model, Case (f) appears to have smaller flux discrepancy for low-flux images, because the images reconstructed by this method tend to have smaller noise-like fluctuations that reduce pixel-wise differences.
More importantly, most groups with the identity loss remain nearly flat over different flux scales, whereas the other groups, due to a lack of identity information, are strongly biased towards having larger flux gaps with increasing flux.

\subsubsection{Local Fluctuation Patterns}

The behavior of noise is characterized as local fluctuation patterns, which can be captured by high-pass filters and displayed in the Fourier space. For evaluation, we convolve the images with the filter shown in Figure~\ref{fig:architecture}(g) and apply a 2D Fourier Transform, same as the operations we perform in the Discriminators. Since the images from the same domain have similar Fourier modes, we stack all the 2D Fourier amplitude maps for the original images and the reconstructed images in each case, respectively.
In Figure~\ref{fig:ft}, we present the Fourier amplitude maps for the $r$-band images. We do not make such comparison for the CFHT images in Cases (g) and (h), since the regridded images used in those cases may have distinct noise behavior from the non-regridded images.
The Fourier map reconstructed by our model resembles the original one, while those obtained by other methods are distinctive. Notably, there are dramatically high peaks located at the center for Cases (e), (g) and (h) in which the Deconvolutional layers or the Pixel Shuffle units are used. These methods overly concentrate on a few high-frequency Fourier modes and thus produce regular ripples shown in Figure~\ref{fig:comparisons}. Although failed regeneration of noise may not necessarily imply inability to learn noise properties, we suggest that only by adding random seeds can we achieve to convert learned noise properties into realistic noise. Finally, Case (f) is comparable to our model for SDSS images, but fails for CFHT images, implying that the Noise Emulators cannot be properly trained due to the negative effect from the Generators, as illustrated in Section~\ref{sec:model}.

\section{Conclusion}

We develop a semi-supervised noise-reconstructed GAN approach to conduct image-to-image translation between two sky surveys. As demonstrated by our experiments on galaxy images, we achieve to learn and reconstruct high-frequency noise using noise emulating modules. We emphasize that this is a noise reconstruction approach rather than a de-noising method as those developed by other work. We also show that a small amount of paired data can greatly alleviate the difficulty in recovering galaxy shapes from the intermediate output inside a translation cycle, suggesting the necessity of paired data even when unpaired data is plentifully available. 

This work is the first step towards investing two-way noise-reconstructed image translation methods in astrophysical studies, which would become a promising image simulation approach complementary to traditional methods (e.g., Markov Chain Monte Carlo). The noise reconstruction techniques might also be applicable in areas other than astrophysics (e.g., sonar imaging \cite{sonar}). As high-fidelity data is a stringent demand in real astrophysical applications, future work will need to focus on not only reconstructing high-quality images but also recovering correct physical properties of the targeted source. We caution that merely minimizing the MSE (Eq.~\ref{eq:id} and Eq.~\ref{eq:cyc}) might be insufficient to preserve salient information relevant to real tasks of various kinds (e.g., \cite{1805.08841}). Therefore, it would be interesting to extend this work in the context of certain astrophysical applications.

\bibliographystyle{IEEEtran}
\bibliography{bib}

\end{document}